\documentclass[conference]{IEEEtran}
\IEEEoverridecommandlockouts
\usepackage{cite}
\usepackage{amsmath,amssymb,amsfonts}
\usepackage{algorithmic}
\usepackage{graphicx}
\usepackage{textcomp}
\usepackage{xcolor}
\usepackage[colorlinks,
            linkcolor=blue,
            anchorcolor=blue,
            citecolor=blue]{hyperref}
\usepackage{stfloats}
\usepackage{booktabs}
\def\BibTeX{{\rm B\kern-.05em{\sc i\kern-.025em b}\kern-.08em
    T\kern-.1667em\lower.7ex\hbox{E}\kern-.125emX}}
\begin{document}

\title{SAM on Medical Images: A Comprehensive Study on Three Prompt Modes\\
\thanks{This study was supported by National Key Research and Development Program of China (2020YFB1711500, 2020YFB1711503), and the 1·3·5 project for disciplines of excellence, West China Hospital, Sichuan University (ZYYC21004).}
}


\author{
\IEEEauthorblockN{
Dongjie Cheng\IEEEauthorrefmark{2},
Ziyuan Qin\IEEEauthorrefmark{2},
Zekun Jiang\IEEEauthorrefmark{2},
Shaoting Zhang\IEEEauthorrefmark{3},
Qicheng Lao\IEEEauthorrefmark{4}\IEEEauthorrefmark{7},
Kang Li\IEEEauthorrefmark{2}\IEEEauthorrefmark{7},
}
\IEEEauthorblockA{\IEEEauthorrefmark{2}West China Biomedical Big Data Center,West China Hospital, Sichuan University, Chengdu, China}
\IEEEauthorblockA{\IEEEauthorrefmark{3}Shanghai Artificial Intelligence Laboratory, Shanghai, China}
\IEEEauthorblockA{\IEEEauthorrefmark{4}School of Artificial Intelligence, BUPT, Beijing, China}
\IEEEauthorblockA{\IEEEauthorrefmark{7}Corresponding Author: Qicheng Lao, Kang Li \quad Email: qicheng.lao@bupt.edu.cn, likang@wchscu.cn}}


\maketitle

\begin{abstract}
The Segment Anything Model (SAM) made an eye-catching debut recently and inspired many researchers to explore its potential and limitation in terms of zero-shot generalization capability. As the first promptable foundation model for segmentation tasks, it was trained on a large dataset with an unprecedented number of images and annotations. This large-scale dataset and its promptable nature endow the model with strong zero-shot generalization. Although the SAM has shown competitive performance on several datasets, we still want to investigate its zero-shot generalization on medical images. As we know, the acquisition of medical image annotation usually requires a lot of effort from professional practitioners. Therefore, if there exists a foundation model that can give high-quality mask prediction simply based on a few point prompts, this model will undoubtedly become the game changer for medical image analysis. To evaluate whether SAM has the potential to become the foundation model for medical image segmentation tasks, we collected more than 12 public medical image datasets that cover various organs and modalities. We also explore what kind of prompt can lead to the best zero-shot performance with different modalities. Furthermore, we find that a pattern shows that the perturbation of the box size will significantly change the prediction accuracy. 
Finally, Extensive experiments show that the predicted mask quality varied a lot among different datasets. And providing proper prompts, such as bounding boxes, to the SAM will significantly increase its performance.
\end{abstract}


\section{Introduction}
Recently, large language models pre-trained with billions or even trillions of parameters have become the most influential trends in AI. Large language models (LLM) such as ChatGPT or LLaMA can be easily extended to different tasks and scenarios without any further training or tuning. All you need is to prompt the LLMs to activate the relevant knowledge. These amazing zero-shot generalizations make the name "foundation model" become well-known. However, most of the breakthroughs in foundation models happened in either NLP fields or cross-modality tasks such as image text retrieval. However, for many computer vision tasks, such as segmentation, abundant training data does not exist. Thus, the first segmentation foundation model, SAM~\cite{kirillov2023segment}, was born and showed impressive zero-shot inference performance in different domains. On the other hand, segmentation has long been an important topic in medical image analysis, involving the segmentation of organs, abnormalities, tumors, and others. Nevertheless, since domain discrepancy and modality differences exist in medical image tasks, current solutions usually rely on designing a specialized model architecture and training with significant data. Therefore, we need a foundation model that only needs a few prompts to generalize to a new domain or task in medical image analysis. However, acquiring web-scale images and annotations is not realistic. So, in this work, we want to investigate whether the SAM can be the foundation model for medical image analysis segmentation tasks.  

For this purpose, we split our experiments into 3 parts. Firstly, we evaluate the SAM with auto-prompt mode, and conclude that SAM's zero-shot generalization in auto-prompt mode is not enough to compete with the baseline models. We further explore the box-prompt mode and point-prompt mode of the SAM model to investigate their zero-shot generalization. We conduct extensive experiments to show that box-prompt mode achieves higher dice accuracy than other prompt modes. We also observe a pattern that adding jitters to the boxes can significantly affect prediction accuracy. For point-prompt mode, we evaluate with single point, 3 point, and 10 points setting. We conclude that with the increasing of prompt points, the zero-shot performance is approaching that of the box-prompt mode. 

Though SAM achieves competitive zero-shot generalization on many domains and modalities when we give the model proper prompts, the overall performance is still not as good as the trained model with the supervised method. Afterall, SAM provides a reference for the development of medical foundation models, which is worth exploring in more depth in future research.

\begin{figure*}[ht]
\centerline{\includegraphics[width=1.2\linewidth]{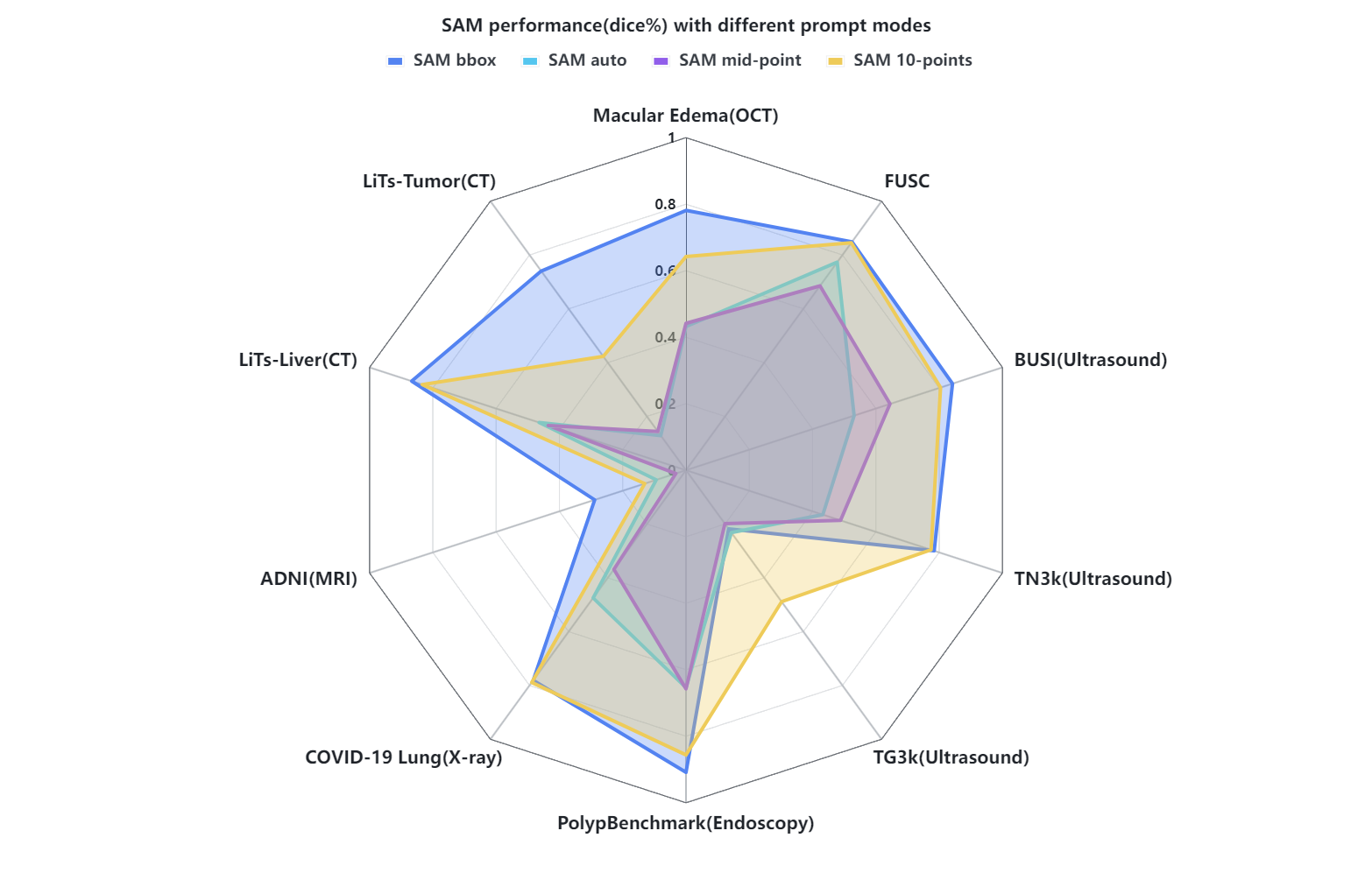}}
\caption{Segmentation results with different prompt modes}
\label{fig1}
\end{figure*}

\section{Related Works}
It has been over three weeks since the publication of "Segment Anything"~\cite{kirillov2023segment}, and a search of SAM-related research on arVix now yields more than 20 preprint articles. We anticipate that the number of research studies using SAM will continue to grow in the coming period, as it has successfully captured the attention of scholars in AI, computer vision (CV), image processing, and various cross-disciplinary application fields.
\subsection{About Segment Anything}
The emergence of LLM has led to a shift in the research goals of many AI researchers toward the development of large or foundation models. This trend has gained immense popularity and is currently one of the hottest topics in AI research. One viewpoint in the current academic community is that emergence capability can be achieved when the model parameters reach a sufficiently large level, resulting in impressive intelligent processing capabilities. As a result, many AI researchers are considering whether the same is true in the CV domain and looking forward to developing CV foundation models. In this context, segmenting anything becomes an imperative foundation model. Recently, two major parallel works were opened, one is SAM~\cite{kirillov2023segment} and the other is "Segment Everything Everywhere All at Once" named SEEM~\cite{zou2023segment}. During the past two weeks, SAM had a greater degree of impact, but SEEM showed that it has many aspects of performance that outperform SAM and also deserve our attention, measurement, and secondary development.

\subsection{SAM in Medical Imaging}
Overall, SAM has opened up the foundation models to facilitate zero-shot processing and secondary development for various downstream tasks. In the field of medical image processing, there are currently more than 10 preprint papers. Seven of these works were focused solely on segmentation in specific medical areas, including digital pathology images~\cite{deng2023segment}, multi-phase liver tumors in CT~\cite{hu2023sam}, abdominal CT~\cite{roy2023sam}, brain MRI~\cite{mohapatra2023brain}, brain tumors in BraTS~\cite{putz2023segment}, polyps~\cite{zhou2023can}, and ophthalmology~\cite{qiu2023learnable}. Meanwhile, three other works, mainly from Harvard Medical School, Duke University, and the University of Toronto, respectively, conducted more comprehensive evaluations of segmentation comparisons across multiple organ sites and medical imaging multi-modalities~\cite{he2023accuracy,mazurowski2023segment,ma2023segment}. Of these, Ma et al~\cite{ma2023segment} performed the most extensively evaluated and provided a tutorial for fine-tuning SAM. Unlike these three work, we focus on which prompt=mode is suitable for the SAM model in medical image analysis. There is also a study that integrates SAM into 3D Slicer~\cite{liu2023samm}, a widely-used open-source medical image processing software. Additionally, a blog called "SAM in Medical Imaging" provides a program to directly process medical image formats using SAM.

\subsection{Other SAM Works}
In other areas, there also has been a considerable amount of research work based on SAM. There are approximately more than 10 studies, of which 2 have evaluated SAM on a broader range of natural images~\cite{ji2023segment,tang2023can}, while the other works have improved upon SAM to varying degrees or used it for other tasks. SAM-Adapter was developed for better segmentation~\cite{chen2023sam}. Moreover, SAM was also used for inpainting anything\cite{yu2023inpaint}, 3D reconstruction\cite{shen2023anything}, style transformer\cite{liu2023any}, counting anything\cite{ma2023sam}, and segmenting 3D anything\cite{cen2023segment}, etc.

\begin{figure*}[ht]
\centerline{\includegraphics[width=1\linewidth]{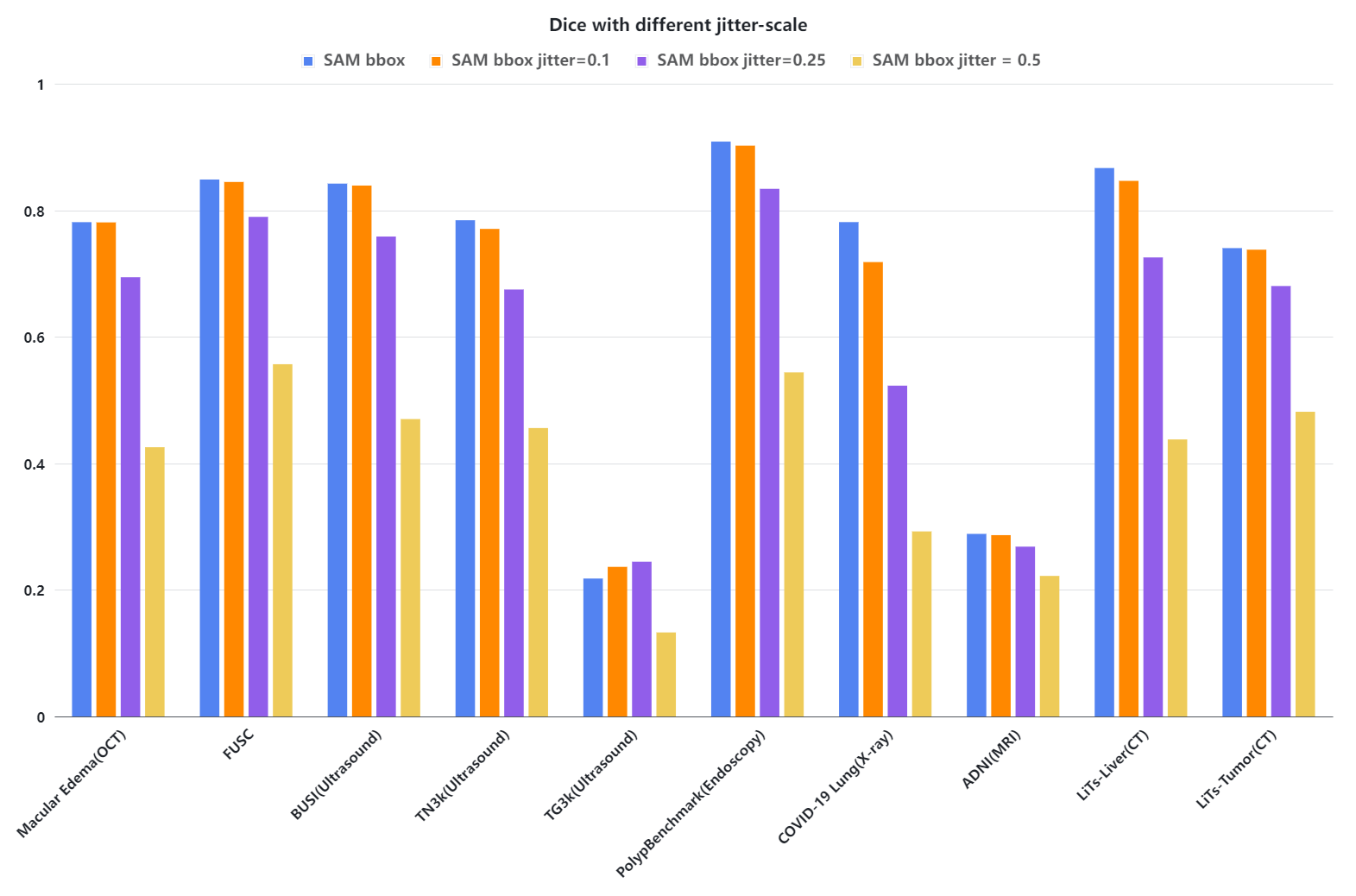}}
\caption{We show the pattern of dice changing with different level of jitter over most of the datasets.}
\label{fig2}
\end{figure*}

\begin{table*}[ht]\tiny
\caption{Segmentation performance comparison in different bbox jitters} 
\label{tab1}
\vspace{-20pt}
\begin{center}
\resizebox{\linewidth}{!}
{
\begin{tabular}{ccccccccccc}
\toprule
Dataset & OCT & FUSC & BUSI & TN3k & TG3k & PolypBenchmark & X-ray & ADNI & LiTs-Liver & LiTs-Tumor\\

\midrule

SAM bbox  & \textbf{0.7812} & 0.8486 & 0.8421 & \textbf{0.7843} & 0.2179 &\textbf{0.9086} &0.7813 & 0.2885 &\textbf{0.8667} &\textbf{0.7402}\\

SAM bbox jitter=0.01  & 0.7807 & 0.8489 & \textbf{0.8424} & 0.7838 & 0.2177  &0.9080 &\textbf{0.7815} &\textbf{0.2887} &0.8665 &0.7398\\

SAM bbox jitter=0.05  & 0.7804 & \textbf{0.8503} & 0.8401 & 0.7821 & 0.2220 &0.9028 &0.7653 &0.2883 &0.8642 &0.7393\\

SAM bbox jitter=0.1  & 0.7808  & 0.8447 & 0.8391 & 0.7705 & 0.2364 &0.9022 &0.7181 &0.2865 &0.8465 & 0.7377\\

SAM bbox jitter=0.25  & 0.6941  & 0.7896 & 0.7585 & 0.6747 & \textbf{0.2446} &0.8339 &0.5228 &0.2684 &0.7255 &0.6803\\

SAM bbox jitter=0.5  & 0.4255 & 0.5567 & 0.4700 & 0.4558 & 0.1328 &0.5438 &0.2924 &0.2221 &0.4378 &0.4816\\

SOTA & \textbf{0.8447} & \textbf{0.8923} & \textbf{0.8577} & \textbf{0.8126} & / &0.8946 &\textbf{0.9840} &/ &\textbf{0.9630} &0.6740\\
\bottomrule
\end{tabular}
}
\end{center}
\end{table*}

\begin{table*}[ht]\tiny
\caption{Segmentation performance comparison in different point prompts} 
\label{tab2}
\vspace{-20pt}
\begin{center}
\resizebox{\linewidth}{!}
{
\begin{tabular}{ccccccccccc}
\toprule
Dataset & OCT & FUSC & BUSI & TN3k & TG3k & PolypBenchmark & X-ray & ADNI & LiTs-Liver & LiTs-Tumor\\

\midrule
SAM Auto  & 0.4318 & 0.7731 & 0.5312  & 0.4324 & 0.2318 & 0.6540 &0.4749 &0.0941 &0.4641 & 0.1288\\

SAM midpt prompted  & 0.4416 & 0.6852 & 0.6447 & 0.4884 & 0.1989 &0.6572 &0.3686 &0.0320 &0.4354 & 0.1442\\

SAM 3pts prompted & 0.4728 & 0.6708 & 0.6180 & 0.6008 & 0.2541 &0.7686 &0.5288 &0.1014 &0.6362 &0.2613\\

SAM 10pts prompted &\textbf{0.6425} & \textbf{0.8453} & \textbf{0.8044} & \textbf{0.7745} & \textbf{0.4889} & \textbf{0.8560} &\textbf{0.7890} & \textbf{0.1303} &\textbf{0.8336} &\textbf{0.4231}\\

SOTA  & \textbf{0.8447} & \textbf{0.8923} & \textbf{0.8577} & \textbf{0.8126} & / &\textbf{0.8946} &\textbf{0.9840} &/ &\textbf{0.9630} &\textbf{0.6740}\\
\bottomrule
\end{tabular}
}
\end{center}
\end{table*}

\begin{figure*}[ht]
\centerline{\includegraphics[width=1\linewidth]{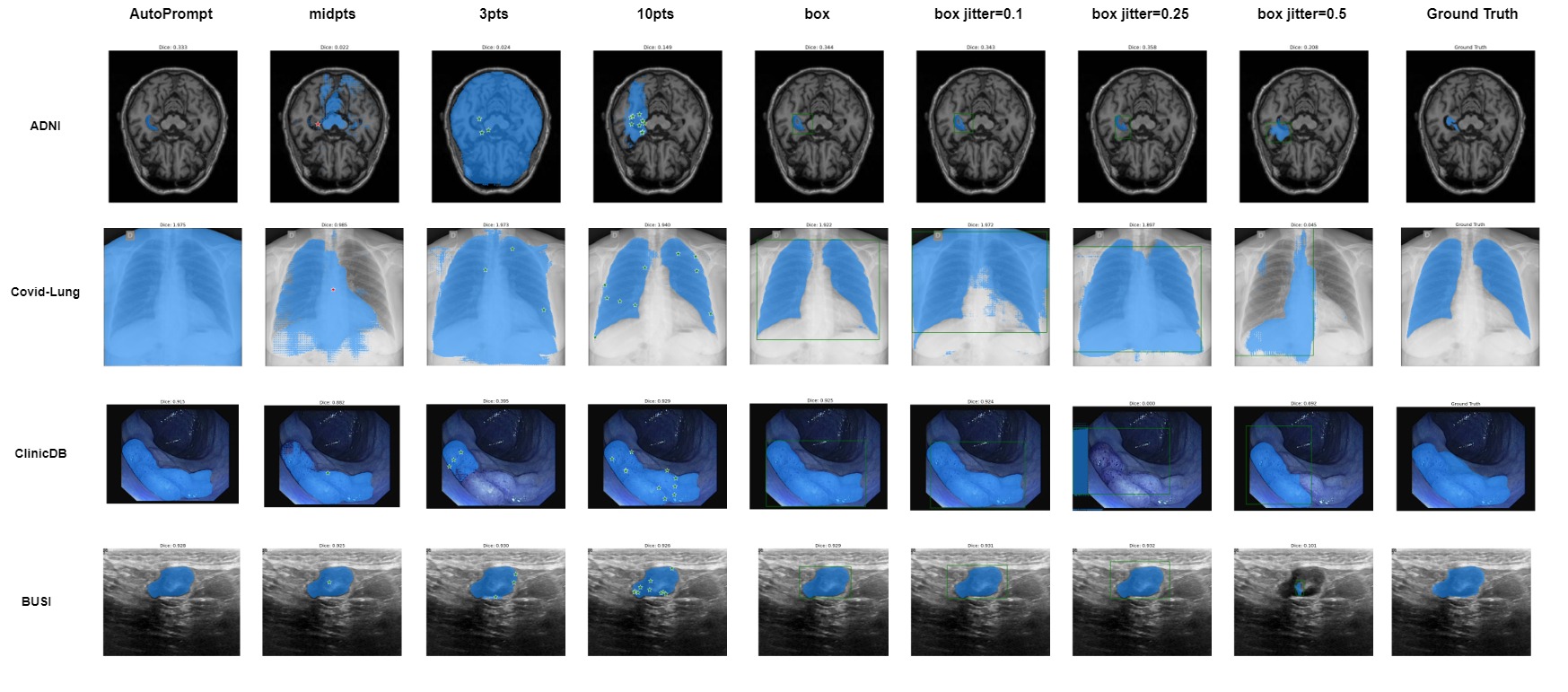}}
\caption{Segmentation results with different prompt modes}
\label{fig3}
\end{figure*}

\section{Methods}
\subsection{Dataset}\label{AA}
We collected 12 different public available datasets covering CT, X-ray, MRI, Endoscopy, Ultrasound, and OCT. (1) \textbf{BUSI}:~\cite{al2020dataset} collected at baseline include breast ultrasound images among women in ages between 25 and 75 years old. (2) \textbf{CVC-300~\cite{cvc-300}, CVC-ClinicDB~\cite{cvc-clinicdb}, CVC-ColonDB~\cite{cvc-colondb}, Kvasir~\cite{kvasir}, and ETIS~\cite{etis}}: These five datasets are part of the polyp region detection benchmark that was used in PraNet~\cite{polpy}. (3) \textbf{FUSC} is from Foot Ulcer Segmentation Challenge 2021~\cite{wang2022fuseg}, with the purpose to segment ulcer on the feet. (4) \textbf{COVID-Lung} is one of the largest benchmark dataset designed for COVID-19 detection and segmentation, with 33,920 CXR images, including 11,956 COVID-19 samples~\cite{TAHIR2021105002}, (5) \textbf{ADNI} is a MRI dataset initiated for Alzheimer's Disease classification~\cite{ADNI}, but it is also used for hippocampus segmentation purpose.  
(6) \textbf{TN3K} and \textbf{TG3K} are from the same dataset but with different segmentation annotations. This dataset containing more than three thousand ultrasound images with high-quality annotation of thyroid nodule and thyroid gland respectively~\cite{GONG2023106389}. (7) \textbf{LiTs} is a data set for the Liver and Liver Tumor Segmentation Benchmark (LiTS)~\cite{bilic2023liver}. (8) \textbf{OCT} dataset is from a well-known Chinese AI challenge platform, initiated by Chinese enterprise~\cite{hu2019automated}.

In the study, we used SAM to perform segmentation and dice computation directly on the test sets of different datasets. Then, the segmentation results were compared with the state-of-the-art (SOTA) results of supervised learning from the related dataset studies. The relevant SOTA results can be found in ~\cite{wang2022fuseg,hu2019automated,kuzinkovas2022detection}, etc.

\subsection{Applying SAM with different prompt modes}
In this work, we test three different prompt modes and thoroughly evaluate their performance on medical image segmentation tasks. As mentioned, the SAM can run in three different modes: auto-prompt, point-prompt, and box-prompt. For the auto-prompt mode, SAM will be automatically prompted with a regular grid of points and predicate a set of masks for each point prompt. For box-prompt mode, we first generate a bounding box according to the ground truth for each object and input the curated bounding boxes as prompts to the SAM with different scales of jitter.  Finally, we also tried different settings for point-prompt mode. Under the single-point prompt setting, we chose the center point of the curated bounding box as our prompt. Unlike most of the evaluation papers for the SAM model, we didn't use the center point of ground truth mask as a prompt because sometimes the center point of an irregular shape will drop outside of the circumscribed area, and we think that will give the SAM model a lousy prompt. However, for some non-contiguous masks, that center point can still fall on the background are that may mislead the SAM model. This conjecture also aligns with the observation mentioned in ~\cite{mazurowski2023segment}. We also test the point-prompt mode with 3 and 10 points settings. 

No matter under which prompt setting, the zero-shot prediction accuracy of the SAM model is still lower than that of the commonly used segmentation models trained with fully supervised methods. Another topic worth diving into is how to generate high-quality prompts, as prompts curated from ground truth mask is no longer available in real-world usage. 

\subsection{How to properly prompt the SAM with boxes and points}
One interesting pattern we observed in the prediction accuracy is decreasing with the increase of the jitter scale. We conjecture that the bounding box scopes more background or unrelated areas after we add more perturbation and jitter. Thus, we can also conclude that the SAM model's prediction quality is sensitive to the box size and its reliability. How to produce the reliable and suitable size of box-prompts will be a possible research topic for applying the SAM model to different domains. For the point-prompt mode, We found that the predicted mask quality is rising and competing with the box-prompt performance with the increase of given prompt points. This pattern is indeed consistent with our intuition. The reason that the box-prompt mode beat the few-point-prompt mode in our experiments could be complicated. One of the contributing factors could be the uneven boundary of some abnormalities. For those not well-circumscribed areas, several points sampled from the mask area can not represent the irregular shape well.

\section{Results}
As shown in Table \ref{tab1}, \ref{tab2}, and Fig \ref{fig1}, the performance of SAM varies among different datasets and prompt modes. Specifically, the box-prompt mode with zero jitters is the best way to utilize the SAM model on medical image tasks. The box-prompt mode achieves the highest average dice compared with the other two modes. This prompt method also occasionally gives results that are close to the SOTA methods. Thus we argue that box-prompt mode might be the most suitable prompt method to leverage the SAM model in medical image segmentation tasks. In General, the Dice of the SAM model with box-prompt mode is 0.

In Fig \ref{fig2}, we show how the prediction accuracy would change after more jitters are applied in all datasets .one can tell that the jitters added to the boxes can significantly affect the prediction mask accuracy. We discussed this observation before. We also give a visualization of some segmentation results in Fig \ref{fig3}.

Overall, the SAM zero-shot results are generally lower than SOTA results, but there are still several datasets that obtain performance exceeding SOTA. There is variability in the results of SAM segmentation on different medical image datasets, and the reasons for this variability and how to make improvements will be further explored in our future studies.

\bibliographystyle{IEEEtrans}
\bibliography{IEEEexample}
\end{document}